# Advanced Intelligent Optimization Algorithms for Multi-Objective Optimal Power Flow in Future Power Systems: A Review

Yuyan Li [1]

[1] School of Electrical and Electronics Engineering, North China Electric Power University, Beijing 102206, China

**Abstract**: This review explores the application of intelligent optimization algorithms to Multi-Objective Optimal Power Flow (MOPF) in enhancing modern power systems. It delves into the challenges posed by the integration of renewables, smart grids, and increasing energy demands, focusing on evolutionary algorithms, swarm intelligence, and deep reinforcement learning. The effectiveness, scalability, and application of these algorithms are analyzed, with findings suggesting that algorithm selection is contingent on the specific MOPF problem at hand, and hybrid approaches offer significant promise. The importance of standard test systems for verifying solutions and the role of software tools in facilitating analysis are emphasized. Future research is directed towards exploiting machine learning for dynamic optimization, embracing decentralized energy systems, and adapting to evolving policy frameworks to improve power system efficiency and sustainability. This review aims to advance MOPF research by highlighting state-of-the-art methodologies and encouraging the development of innovative solutions for future energy challenges.

## 1. Introduction

### 1.1 Background and Significance of the Study

The evolving landscape of global energy systems underscores a pressing need for efficient, reliable, and sustainable power management strategies. Optimal Power Flow (OPF) has long been at the heart of power system engineering, facilitating the optimal allocation of resources within electrical grids to minimize costs while maintaining system security and reliability [1-2]. However, the integration of renewable energy sources, the advent of smart grid technologies, and the increasing demand for electricity have introduced new dimensions of complexity and multi-objectivity into the OPF problem [3-4].

Multi-Objective Optimal Power Flow (MOPF) extends the traditional OPF framework by simultaneously optimizing multiple conflicting objectives, such as minimizing generation cost, minimizing emissions, and maximizing system reliability [5-7]. The significance of MOPF lies in its ability to provide a balanced solution to the trade-offs inherent in modern power systems, accommodating the diverse priorities of stakeholders ranging from utility companies to environmental groups.

The application of intelligent optimization algorithms in solving MOPF problems has gained considerable attention [8-9]. These algorithms, including evolutionary algorithms, swarm intelligence, and deep reinforcement learning (DRL), offer promising avenues to tackle the complexity and scalability challenges of MOPF by exploring the solution space more efficiently and effectively than traditional methods.

### 1.2 Purpose and Scope of the Review

This review aims to systematically explore the application of intelligent optimization algorithms to MOPF problems, highlighting key methodologies, comparing algorithmic performances, and identifying emerging trends and challenges. By doing so, it seeks to provide a comprehensive resource for researchers and practitioners in the field of power systems engineering, offering insights into the state-of-the-art approaches and future directions for MOPF research.

The scope of this review encompasses a broad spectrum of intelligent optimization algorithms applied to various MOPF problems, including pure AC systems, hybrid AC/DC systems, and integrated energy systems (IES). Furthermore, it delves into the implications of incorporating new energy sources, security and stability constraints, and privacy concerns within the MOPF framework. Through a detailed analysis and comparison of algorithmic approaches, this review also aims to shed light on the efficacy, scalability, and applicability of these algorithms to real-world power systems.

In summarizing the current landscape and outlining future perspectives, this review aspires to contribute to the advancement of MOPF research, facilitating the development of more efficient, resilient, and sustainable power systems in the face of rapidly evolving energy demands and technological innovations.

### 1.3 Research Gaps and Contributions

Despite significant advancements in the field of Multi-Objective Optimal Power Flow (MOPF), several research gaps persist that limit the comprehensive understanding and application of intelligent optimization algorithms within modern power systems. These gaps include:

- **Algorithm Performance in Complex Scenarios**: Current literature often lacks in-depth analysis of how different intelligent optimization algorithms perform under the complex, dynamic scenarios characteristic of modern power systems, especially in hybrid AC/DC systems and integrated energy systems (IES).
- **Comparative Efficacy for Renewable Integration**: There is a gap in comparative studies that elucidate the efficacy of these algorithms specifically for optimizing power flow with high penetrations of renewable energy, considering the stochastic nature of such sources.
- **Adaptation to Policy and Market Changes**: Limited research has been conducted on how intelligent optimization algorithms for MOPF adapt to rapidly evolving policy environments and market mechanisms, particularly in deregulated energy markets.
- **Utilization of Advanced Computing Techniques**: The potential of leveraging advanced computing techniques, including machine learning and blockchain, for enhancing MOPF solutions is not adequately explored, particularly in terms of privacy preservation and decentralized optimization.

In addressing these gaps, our review contributes to the field in the following ways:

- **Detailed Algorithmic Analysis**: We provide a nuanced comparison of intelligent optimization algorithms, focusing on their performance, adaptability, and scalability in the complex environments of modern power systems, including detailed discussions on hybrid AC/DC and IES scenarios.

- **Focus on Renewable Energy Optimization**: This review highlights specific strategies and algorithmic adaptations necessary for efficiently integrating renewable energy sources into power systems, addressing the challenges of variability and uncertainty head-on.
- **Insights into Policy and Regulation**: We explore the implications of energy policies and regulatory frameworks on the selection and effectiveness of MOPF strategies, offering insights into how optimization approaches can remain flexible and effective amidst policy shifts.
- **Exploration of Interdisciplinary Approaches**: By delving into the application of advanced computing techniques—such as machine learning for predictive modeling and blockchain for secure, decentralized energy transactions—our review underscores the untapped potential of these technologies in MOPF research.

Through these contributions, the review aims to fill existing research voids by providing a comprehensive, up-to-date synthesis of knowledge on intelligent optimization algorithms for MOPF, paving the way for future innovations and the development of more robust, efficient, and policy-responsive power system optimization strategies.

## 2. Fundamental Concepts

### 2.1 Optimal Power Flow

#### 2.1.1 Definition and Objectives

OPF is a mathematical optimization model used in electrical power systems to determine the optimal operating conditions that meet specific objectives, such as minimizing generation cost, reducing power losses, or maximizing system security, under a set of physical and operational constraints [10]. These constraints include power balance equations, generator limits, voltage limits, and transmission line capacity limits, ensuring the operational feasibility of the power system.

The primary objective of OPF is to optimize the power system's performance by finding the most economical way to generate and distribute electricity across the network while adhering to all technical constraints and ensuring a stable and reliable supply [11]. By solving the OPF problem, power system operators can make informed decisions on how best to dispatch generation resources, control voltage levels, and manage reactive power, thus optimizing the overall efficiency and reliability of the power grid [12].

#### 2.1.2 Application Contexts

The applications of OPF span various aspects of power system operations and planning, reflecting its fundamental role in modern power systems management:

- **Generation Dispatch**: OPF is used to determine the optimal output of each generating unit in the system so that the total generation cost is minimized while meeting the demand and operating within technical constraints.
- **Congestion Management**: In times of network congestion, OPF can be applied to alleviate congestion by optimizing the power flow patterns across the network, thereby enhancing the system’s operational efficiency and stability.

- **Voltage Control**: OPF solutions can include optimal settings for voltage control devices, such as transformers and capacitors, to maintain voltage levels within the desired range across the power system.
- **Renewable Energy Integration**: As renewable energy sources are increasingly integrated into the power grid, OPF helps in managing the variability and uncertainty associated with these sources, optimizing the mix of conventional and renewable generation to achieve cost-effective and reliable operation.
- **Emission Reduction**: By including environmental objectives, such as minimizing CO2 emissions, OPF can be used to promote cleaner energy production and support the transition towards sustainable power systems.

In summary, OPF plays a crucial role in ensuring the efficient, reliable, and sustainable operation of power systems. It serves as a fundamental tool for addressing the complex challenges posed by modern electrical grids, especially with the integration of renewable energy sources and the advent of smart grid technologies. Through the application of OPF, power system operators can optimize the performance of the grid, ensuring that electricity is generated and distributed in the most efficient and cost-effective manner possible.

## 2.2 Multi-Objective Optimal Power Flow

### 2.2.1 Importance of MOPF

MOPF extends the OPF framework by considering multiple, often conflicting objectives simultaneously [13-14]. This approach is of paramount importance in modern power systems where operators must balance a variety of considerations, such as cost, environmental impact, reliability, and system security [15-17]. MOPF reflects the multi-faceted nature of decision-making in power system operations, acknowledging that optimizing one objective may lead to suboptimal results in others [18-19].

The increasing penetration of renewable energy sources, coupled with the rising demand for electricity and the need for sustainable operations, underscores the significance of MOPF [20]. It provides a more holistic approach to power system optimization, allowing for a balanced consideration of economic, environmental, and technical criteria [21]. By addressing multiple objectives, MOPF facilitates the identification of trade-offs and the selection of solutions that best meet the diverse needs and preferences of stakeholders, from utility operators to consumers and regulatory bodies [22-23].

### 2.2.2 Main Goals and Challenges of MOPF

**1) Main Goals:**

- **Economic Efficiency**: Minimizing the total operational cost, including fuel costs, maintenance, and environmental penalties, while efficiently integrating renewable energy sources [24].
- **Environmental Sustainability**: Reducing the environmental impact of power generation by minimizing emissions and promoting the use of clean and renewable energy sources [25].

- **System Reliability and Security**: Ensuring a reliable and uninterrupted power supply by maintaining system stability, managing congestion, and adhering to security constraints.

**2) Challenges:**

- **Complexity and Computational Demand**: The addition of multiple objectives significantly increases the complexity and computational requirements of the optimization problem, making it challenging to find optimal solutions within reasonable time frames [26].
- **Conflicting Objectives**: Different objectives may conflict with one another (e.g., minimizing cost vs. minimizing emissions), requiring the development of methodologies to identify acceptable trade-offs and compromise solutions [27].
- **Uncertainty and Variability**: The integration of renewable energy sources introduces uncertainty and variability into the power system, complicating the optimization process. MOPF must account for these factors to ensure robust and flexible solutions.
- **Stakeholder Engagement**: Successfully applying MOPF requires considering the preferences and priorities of multiple stakeholders, necessitating effective communication and negotiation to reach consensus on acceptable solutions [28].
- **Technological and Regulatory Constraints**: Adapting to rapidly evolving technologies and regulatory environments poses additional challenges for MOPF, requiring continuous updates and adaptations of optimization models and approaches.

To summarize, MOPF represents a critical advancement in the optimization of power systems, addressing the complex and multi-dimensional challenges of modern electrical grids. It offers a framework for making more informed and balanced decisions, though it also introduces significant computational and methodological challenges that require ongoing research and innovation.

## 3. Application of Intelligent Optimization Algorithms in MOPF

### 3.1 Pure AC Systems

The application of intelligent optimization algorithms to solve MOPF problems in pure AC systems has become increasingly popular due to their ability to handle complex, nonlinear, and multi-modal optimization landscapes. These algorithms offer promising solutions where traditional methods may struggle, especially in the context of multi-objective and high-dimensional problems prevalent in AC power systems. Below, we explore various types of algorithms applied to MOPF in pure AC systems, along with application examples and performance comparisons.

#### 3.1.1 Types of Algorithms and Application Examples

**1) Evolutionary Algorithms (EAs)**: EAs, including Genetic Algorithms (GA) and Evolutionary Programming (EP), are widely used for MOPF problems due to their robustness in handling multiple objectives and exploring large search spaces [29-30]. For instance, a GA can be employed to minimize both the fuel cost and emission levels in power generation simultaneously, offering a set of Pareto-optimal solutions that highlight the trade-off between economic efficiency and environmental sustainability [31]. The flowchart of Genetic Algorithm is shown in Fig. 1.

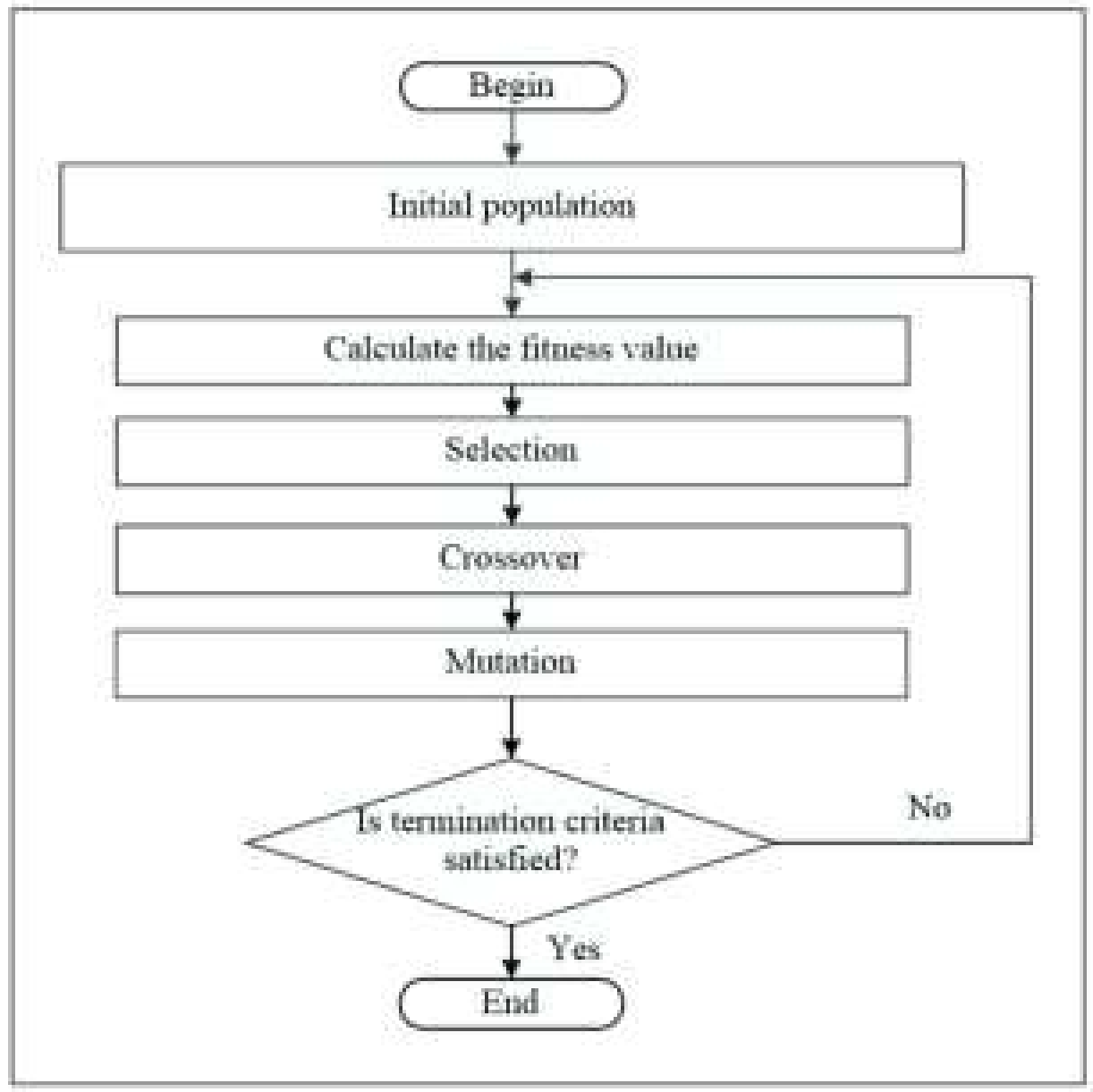

Fig. 1 Flowchart of Genetic Algorithm

**2) Particle Swarm Optimization (PSO)**: PSO is another popular choice for MOPF problems, favored for its simplicity and effectiveness. An application example includes optimizing the cost and voltage stability margin in an AC system, where PSO's ability to converge quickly to high-quality solutions is particularly beneficial [32].

PSO is a population-based optimization algorithm inspired by the social behavior of birds flocking or fish schooling [33]. In PSO, each solution within the search space is represented as a "particle," and the entire population is referred to as a "swarm." Each particle adjusts its position in the search space based on its own experience as well as the experiences of neighboring particles, aiming to find the optimum solution. The core of the PSO algorithm lies in the update equations for velocity and position of the particles [34-35], which are as follows:

$$v_i^{(t+1)} = w \cdot v_i^{(t)} + c_1 \cdot rand() \cdot (pbest_i - x_i^{(t)}) + c_2 \cdot Rand() \cdot (gbest - x_i^{(t)}) \quad (1)$$

$$x_i^{(t+1)} = x_i^{(t)} + v_i^{(t+1)} \quad (2)$$

Where, $v_i^{(t)}$ and $x_i^{(t)}$ are the velocity and position of particle $i$ at time $t$, $pbest_i$ represents the best position particle $i$ has found so far (personal best). $w$ is the inertia weight, balancing the exploration and exploitation capabilities of the swarm. $c_1$ and $c_2$ are cognitive and social parameters, respectively, representing the tendency of particles to follow their own best position and the swarm's best position. $rand()$ and $Rand()$ are random numbers between [0,1], providing stochasticity to the algorithm.

**3) Differential Evolution (DE)**: DE has been applied to MOPF for its strong convergence properties and ease of implementation [36]. An example application is the simultaneous optimization of power loss and load balancing in an AC system, where DE demonstrates superior performance in finding optimal solutions efficiently [37].

**4) Artificial Neural Networks (ANNs) and Deep Learning (DL):** While not optimization algorithms per se, ANNs and deep learning models have been employed to model the complex relationships in power systems and predict optimal power flows [38]. These models can be integrated with optimization algorithms to enhance their performance, especially in dynamic and uncertain environments [39].

### 3.1.2 Performance Comparison and Evaluation

The performance of intelligent optimization algorithms in MOPF problems is typically evaluated based on criteria such as solution quality, computational efficiency, convergence rate, and the ability to find diverse sets of Pareto-optimal solutions. Comparative studies have shown that no single algorithm outperforms others in all scenarios due to the varying nature of MOPF problems and their specific constraints and objectives. For example:

- **Solution Quality and Diversity**: Evolutionary Algorithms, such as GA and Multi-Objective Evolutionary Algorithms (MOEAs), often excel in discovering a diverse set of Pareto-optimal solutions, providing decision-makers with a range of trade-off options. However, they may require larger computational resources and longer run times.
- **Computational Efficiency**: PSO and DE are generally more computationally efficient and faster to converge than EAs, making them suitable for scenarios where computational resources are limited or real-time optimization is required.
- **Robustness and Adaptability**: Algorithms like DE and PSO exhibit robust performance across various MOPF problems, but their effectiveness can be further enhanced through hybridization with other techniques, such as local search methods or machine learning models, to improve solution accuracy and adaptability to dynamic conditions.

### 3.1.3 Challenges and Limitations in MOPF Algorithms

While multi-objective optimization frameworks offer substantial benefits for power system operations, their practical deployment is not without challenges. This section details the limitations inherent in these methods and the obstacles encountered during their implementation. For example, while GAs are praised for their robustness and ability to find global optima, they often suffer from slow convergence rates and can be computationally expensive, limiting their use in real-time applications. Similarly, PSO methods can quickly converge to a solution but are prone to getting trapped in local optima [40,41], which may not be ideal in complex power grid scenarios.

Furthermore, the complexity of implementing these algorithms in a real-world setting often requires a high level of customization and tuning to adapt to specific grid characteristics, which can be resource-intensive. The integration of these algorithms into existing power system infrastructures also poses significant technical and regulatory challenges. Additionally, the lack of standardized benchmarks for comparing different MOPF methods can make it difficult for practitioners to select the most appropriate tool for their needs.

In summary, the selection of an intelligent optimization algorithm for MOPF in pure AC systems depends on the specific requirements of the problem, including the objectives to be optimized, the system's complexity, and the computational resources available. Hybrid approaches and continuous advancements in algorithmic strategies offer promising pathways to addressing the multifaceted challenges of MOPF in pure AC systems.

### 3.2 Hybrid AC/DC Systems

The integration of AC and DC systems into hybrid AC/DC power systems presents unique opportunities and challenges for the optimization of power flow [42]. These hybrid systems combine the advantages of both AC and DC, such as the efficient long-distance transmission capabilities of DC and the widespread infrastructure and technology base of AC. Voltage Source Converter-based High Voltage Direct Current (VSC-HVDC) technology is becoming increasingly pivotal in the electric industry's evolution [43-45]. This innovative HVDC technology offers notable improvements over traditional Line-Commuted Converter-based HVDC (LCC-HVDC) systems. Key benefits of VSC-HVDC include the facilitation of renewable energy integration, the independent control of active and reactive power, and the ability to power passive networks [46-48]. The deployment of numerous significant VSC-HVDC projects has also delivered substantial economic and social advantages. Consequently, solving OPF problem for AC/DC grids is crucial [49-53]. This section explores the characteristics of hybrid AC/DC systems, their optimization needs, and how intelligent optimization algorithms have been applied to address these needs, along with the outcomes of such applications.

#### 3.2.1 Characteristics of Hybrid Systems and Optimization Needs

**1) Characteristics:**

- **Efficiency and Flexibility**: Hybrid systems offer improved transmission efficiency over long distances and greater flexibility in integrating renewable energy sources, thanks to the high efficiency of DC transmission and the ability to connect diverse generation sources [54-56].
- **Complexity in Operation and Control**: The presence of both AC and DC components increases the complexity of system operation and control, requiring advanced management strategies to ensure stability, reliability, and efficiency [57-59].
- **Interconnection Challenges**: Effective interconnection between AC and DC grids involves technical challenges, including synchronization, power conversion, and maintaining power quality and system stability [60-61].

**2) Optimization Needs:**

- **Coordinated Control**: Ensuring optimal power flow across both AC and DC components necessitates coordinated control strategies that can dynamically adjust to changing load conditions and generation capacities [62-63]. For example, the fundamental control strategies for VSC-HVDC are categorized into four main types: 1) maintaining constant DC voltage and reactive power; 2) maintaining constant DC voltage and controlling AC voltage; 3) regulating both active and reactive power, known as constant PQ-control; 4) managing constant active power alongside AC voltage control, referred to as constant PV-control [64-65]. In the context of VSC-Multi-Terminal DC (VSC-MTDC) systems, it is essential for at least one terminal to implement constant DC voltage control to ensure power equilibrium. Other terminals might utilize constant PQ- or PV-control strategies. Additionally, droop control is emerging as a favored strategy for managing VSC-MTDC systems due to its efficiency in balancing system demands [66].
- **Renewable Integration**: Optimizing the integration of variable renewable energy sources (such as wind and solar) into hybrid systems to minimize losses and maximize efficiency.

- **System Stability and Reliability**: Maintaining system stability and reliability in the face of the inherent variability of power flows and the complex dynamics of hybrid systems.
- **Economic Operation**: Achieving cost-effective operation by minimizing losses and operational costs, considering the unique characteristics and requirements of hybrid systems.

### 3.2.2 Problem Formulation

To facilitate discussion without loss of generality, the following discussion uses an AC/DC system incorporating VSC-HVDC as an example to describe the problem formulation of the MOPF model for AC/DC systems. This section discusses a Multi-Objective Optimal Power Flow (MOPF) model for AC/DC systems incorporating Voltage Source Converter-based High Voltage Direct Current (VSC-HVDC). This model aims to balance economic factors, such as voltage deviation, with environmental benefits [67-68]. Key objectives in the model include:

#### 1) Minimizing System Active Power Losses

This is defined by the following equation [66]:

$$P_{\text{loss}}=\sum_{i=1}^{N}U_i\sum_{j=1}^{N}U_j(G_{ij}\cos\theta_{ij}+B_{ij}\sin\theta_{ij})+P_{DC_loss} \tag{3}$$

Here, where $O$ is the total active power losses of AC/DC grids, $N$ is the number of buses in the AC grid, $U_i$ and $U_j$ are respectively the voltage amplitudes of bus $i$ and bus $j$; $G_{ij}$, $B_{ij}$ and $\theta_{ij}$ are respectively the conductance, susceptance and phase-angle difference between buses $i$ and $j$. $P_{DC_loss}$ is the power losses in the DC grid, comprising power losses of converter stations and DC lines. As the main part of $P_{DC_loss}$, the converter station loss $P_{DC_loss}$ is determined by the converter current $I_c$.

#### 2) Emissions of Polluting Gases

$$E=\sum_{i=1}^{N_G}(\alpha_i P_{G,i}^2+\beta_i P_{G,i}+\gamma_i) \tag{4}$$

where $E$ is the amount of polluting gas emissions, $N_G$ is the number of generators, $P_{G,i}$ is the active power output of the $i$th generator, $\alpha_i$, $\beta_i$ and $\gamma_i$ are respectively the factors of polluting gases emissions of generator *i*.

#### 3) Voltage Deviation Index

For improving the voltage quality and operational security of the AC/DC grid, the voltage deviation index $V_{de}$ is used as the objective function [69].

$$V_{de}=\sum_{j=1}^{N}(U_j-U_{set,j})^2+\sum_{k=1}^{N_{dc}}(U_{dc,k}-U_{set,dc,k})^2 \tag{5}$$

where $U_j$ is the voltage of AC bus $j$, and $U_{set,j}$ indicates the predefined value of $U_j$; $U_{dc,k}$ is the voltage of DC bus $k$, and $U_{set,dc,k}$ indicates the predefined value of $U_{dc,k}$.

The constraints for the system include both equality constraints (such as power balance equations) and inequality constraints (like generator limits and voltage bounds). These constraints ensure that the solutions remain feasible and adhere to physical and operational limits of the power systems.

This problem formulation allows for the systematic evaluation of trade-offs between different operational objectives in hybrid AC/DC grids, facilitating more informed and sustainable decision-making in the management of modern power networks.

### 3.2.3 Intelligent Optimization Algorithms Applied and Their Outcomes

**1) Algorithms Applied:**

- **Multi-Objective Evolutionary Algorithms**: MOEAs such as Multi-Objective Particle Swarm Optimization (MOPSO) and Non-dominated Sorting Genetic Algorithm II (NSGA-II) have been effectively applied to optimize multiple conflicting objectives in hybrid AC/DC systems, such as minimizing losses and operational costs while maximizing reliability and stability. For example, Pareto-optimal solutions can be identified using MOPSO, as illustrated in Fig. 2 [69].

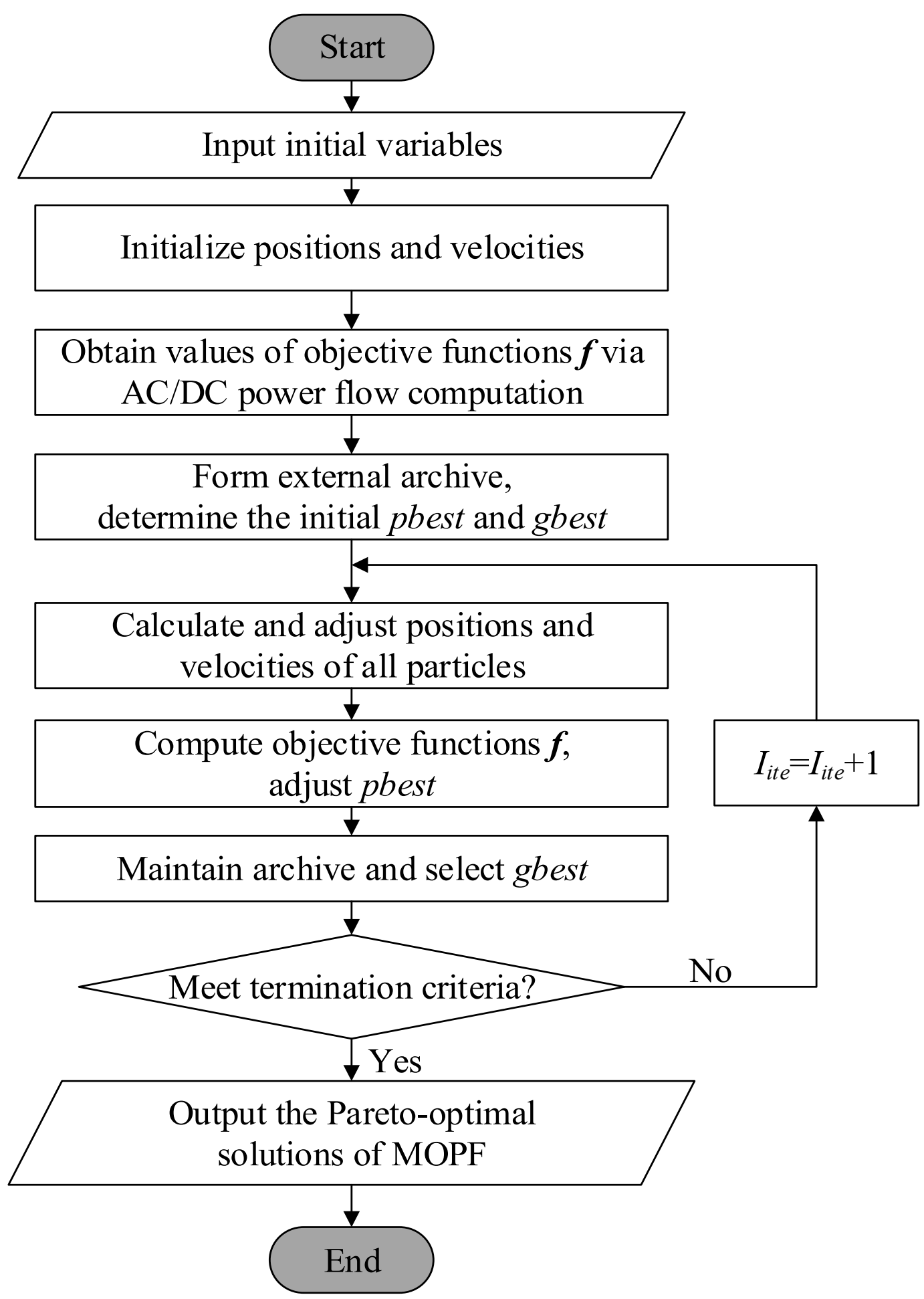

**Fig. 2. Flowchart of optimization scheme using MOPSO**

- **Hybrid Optimization Techniques**: Hybrid optimization techniques combine multiple algorithms to leverage their unique strengths, enhancing solution quality and efficiency for complex, multi-dimensional problem. For instance, integrating PSO with GA can quickly converge to promising areas while diversifying searches to escape local optima. This approach not only improves the robustness and adaptability of solutions, particularly in systems like integrated energy management, but also reduces the need for problem-specific tuning. By dynamically balancing exploration and exploitation, hybrid techniques can effectively address real-world challenges across various domains, offering a versatile tool in advanced optimization scenarios.

**2) Outcomes:**

- **Improved System Efficiency**: Applications of intelligent optimization algorithms have resulted in more efficient power transmission and distribution, with reduced losses and enhanced integration of renewable energy sources.

- **Enhanced Stability and Reliability**: Algorithms have facilitated the development of control strategies that improve the stability and reliability of hybrid systems, even under fluctuating loads and generation capacities.

- **Economic Benefits**: Optimized operation of hybrid AC/DC systems has led to cost savings through improved efficiency, reduced losses, and better management of renewable energy sources.

  For example, when the predefined maximum iterations are reached, the distribution of Pareto-optimal solutions of NSGA-II and MOPSO in the objective function space are shown in Fig. 3 [69].

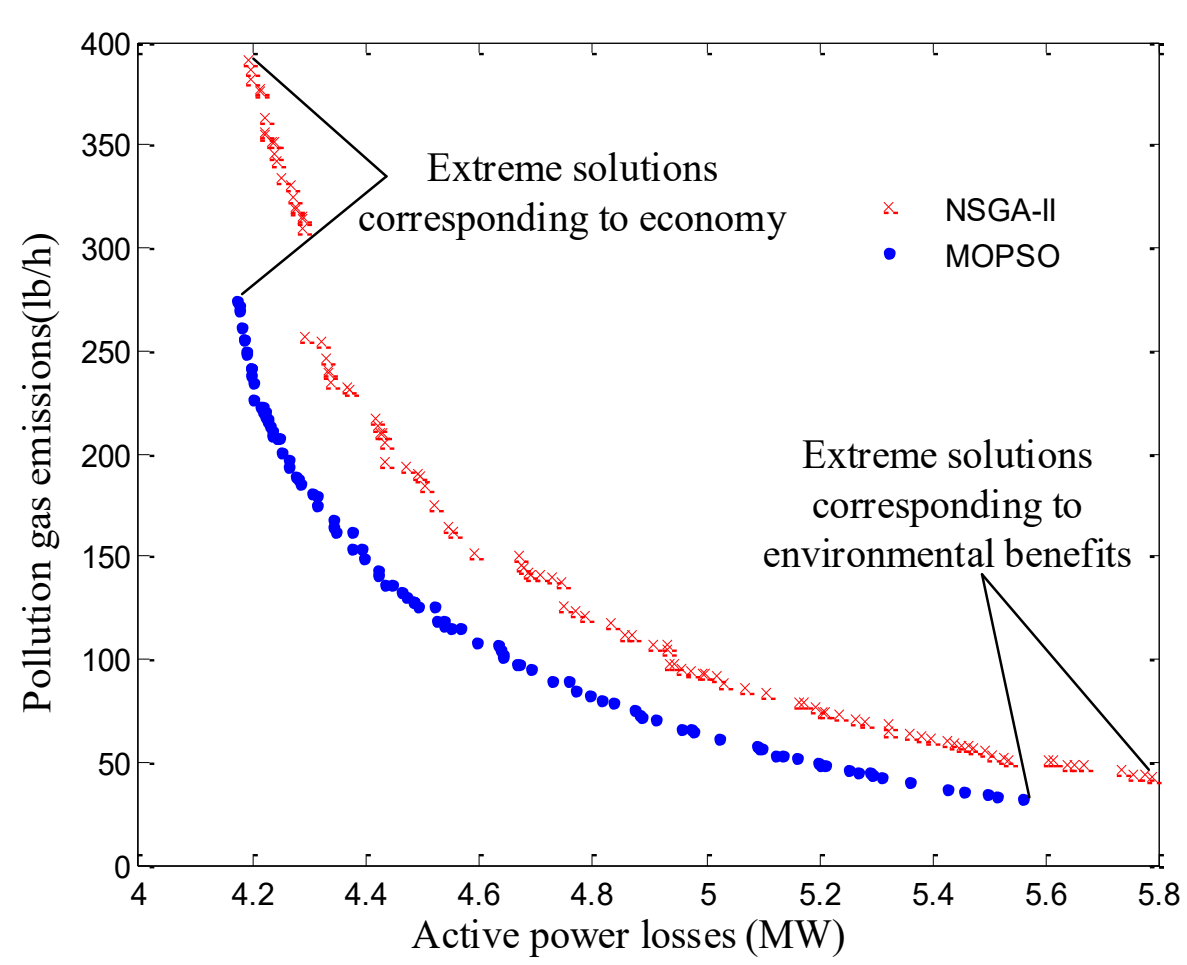

**Fig. 3. Distribution of Pareto frontiers obtained by NSGA-II and MOPSO**

From Fig. 3, it can be seen that: The two objectives, minimize active power loss and emissions of polluting gases, are conflicting to each other. In other words, seeking the minimum of O will inevitably lead to the increase of E at the expense; and vice versa.

Thus, the application of intelligent optimization algorithms to hybrid AC/DC systems addresses the unique challenges and optimization needs of these complex systems. By leveraging the capabilities of these algorithms, it is possible to achieve significant improvements in efficiency, reliability, stability, and economic operation, paving the way for more effective integration and utilization of hybrid power systems in the modern energy landscape.

## 4. Current Issues and Future Perspectives in MOPF

### 4.1 OPF in Hybrid AC/DC Systems

The OPF problem in hybrid AC/DC systems represents a cutting-edge research area that reflects the evolving complexity and needs of modern electrical power systems [70-71]. These hybrid systems, integrating the robustness of AC power grids with the efficiency of DC transmission, are increasingly recognized for their potential to enhance grid capacity, reliability, and facilitate the integration of renewable energy sources [70]. Here, we delve into the current research status and explore future research directions in this dynamic field.

#### 1) Current Research Status

**Integration and Control**: A significant portion of current research focuses on the seamless integration of AC and DC grids, emphasizing the development of advanced control strategies that ensure efficient and reliable power flow between the two systems [72]. This includes the optimization of power converters and the implementation of sophisticated control mechanisms to manage the variability of renewable energy sources and demand fluctuations.

**Modeling and Simulation**: Enhancing the modeling and simulation tools for hybrid AC/DC systems is another active area of research. Accurate models are crucial for analyzing system behavior under various conditions, optimizing the power flow, and assessing the impact of different configurations and control strategies on system performance.

**Renewable Energy Integration**: With the global push towards sustainable energy, optimizing the integration of renewable energy sources into hybrid AC/DC systems is a priority. Research is focused on developing algorithms that can dynamically adjust to the intermittent nature of renewables, ensuring stability and minimizing losses.

**Security and Stability**: Ensuring the security and stability of hybrid AC/DC systems amidst the increasing complexity and potential cyber-physical threats is a critical research area. This includes developing robust security frameworks and stability analysis tools to prevent and mitigate disruptions.

#### 2) Future Research Directions

**Advanced Optimization Algorithms**: There is a continuous need for the development of more advanced, efficient, and scalable optimization algorithms tailored to the unique challenges of hybrid AC/DC systems. This includes leveraging machine learning and artificial intelligence (AI) to enhance the adaptability and efficiency of optimization processes.

**Decentralized and Distributed Control**: As the grid becomes more decentralized with the proliferation of distributed energy resources (DERs), future research will likely focus on decentralized and distributed control strategies for OPF in hybrid systems. This approach can enhance system flexibility, resilience, and facilitate local energy management.

**Energy Storage Integration**: Optimizing the role of energy storage in hybrid AC/DC systems will be crucial for managing variability and enhancing system reliability. Future research will explore innovative strategies for the integration and control of energy storage solutions, including batteries, supercapacitors, and other technologies.

**Regulatory and Market Frameworks**: Adapting regulatory and market frameworks to accommodate the unique features of hybrid AC/DC systems and promote their efficient operation and expansion is a future research area. This includes developing policies and market mechanisms that incentivize the integration of renewables, demand response, and energy storage.

**Interoperability and Standards Development**: Establishing interoperability standards and protocols for hybrid AC/DC systems is essential to ensure compatibility, efficiency, and security across different system components and interfaces. Future research will focus on developing these standards to facilitate system integration and expansion.

To summarize, OPF in hybrid AC/DC systems is poised for significant advancements, driven by the need to integrate renewable energy sources, enhance grid reliability and efficiency, and meet the evolving demands of the energy sector. Addressing the current challenges and exploring new research directions will be crucial for unlocking the full potential of hybrid AC/DC systems in the future energy landscape [73].

### 4.2 OPF Considering Renewable Energy Sources Integration

The integration of renewable energy sources into power systems has become a pivotal aspect of global efforts towards sustainable energy development [74-76]. This shift introduces new dynamics into the OPF problem, necessitating adaptations in strategies and algorithms to effectively manage the inherent characteristics and impacts of these new energy sources [77].

#### 4.2.1 Characteristics of Renewable Energy Sources and Their Impacts on OPF

**Variability and Uncertainty**: Renewable energy sources, such as wind and solar power, are inherently variable and uncertain due to their dependence on weather conditions. This unpredictability introduces challenges in maintaining the balance between supply and demand, affecting the stability and reliability of the power grid [78-80].

**Distributed Generation**: Unlike conventional power plants, renewable energy sources are often distributed across the grid. This decentralization can lead to reverse power flow, voltage fluctuations, and other phenomena that traditional OPF models and control strategies may not be equipped to handle [81-82].

**Low Marginal Cost**: Renewable energy sources have a low marginal cost of production, which can significantly impact the economic dispatch within OPF solutions, favoring renewables over traditional fossil-fuel-based generation [83].

#### 4.2.2 Strategies and Algorithm Improvements

**Stochastic and Robust Optimization**: To address the variability and uncertainty of renewable energy, stochastic and robust optimization methods have been developed. These approaches incorporate uncertainty directly into the OPF model, allowing for solutions that are feasible under a range of possible scenarios, thereby enhancing system resilience [84-87].

**Distributed and Decentralized Algorithms**: Reflecting the distributed nature of renewable energy sources, distributed and decentralized OPF algorithms have gained prominence. These algorithms allow for local optimization at the level of distributed generators and storage devices, facilitating more efficient integration of renewables and enhancing grid flexibility [88-89].

**Multi-Objective Optimization**: The integration of renewable energy sources often requires balancing multiple objectives, such as minimizing cost, emissions, and maintaining system stability [90]. Multi-objective optimization algorithms, including evolutionary algorithms and swarm intelligence, provide a framework for finding compromise solutions that consider the trade-offs between these conflicting objectives [91-92].

**Forecasting and Predictive Control**: Advanced forecasting techniques, utilizing machine learning and big data analytics, improve the predictability of renewable energy output. Integrating these forecasts into predictive control strategies allows OPF to dynamically adjust to anticipated changes in renewable generation, minimizing the impact of variability on grid operation [93].

**Grid-Scale Energy Storage Integration**: Incorporating grid-scale energy storage into OPF models enables the mitigation of renewable variability by storing excess energy during periods of high generation and releasing it during low generation. Optimization algorithms are being developed to manage the operation of storage devices in conjunction with renewable sources, optimizing their combined contribution to the grid [94-95].

This section has demonstrated the integration of renewable energy sources into the power grid introduces significant challenges and opportunities for OPF. Addressing these challenges requires innovative strategies and improvements in optimization algorithms. Future research and development in this area will continue to focus on enhancing the flexibility, reliability, and sustainability of power systems, facilitating a smoother transition to a renewable energy-dominated future.

### 4.3 OPF Considering Security and Stability Constraints

The inclusion of security and stability constraints in OPF models is critical for ensuring the reliable operation of power systems, especially in the face of increasing complexity and the integration of renewable energy sources [96]. These constraints address the physical and operational limits of power systems to prevent failures and ensure resilience against disturbances [97].

#### 4.3.1 Types and Importance of Stability Constraints

**Voltage Stability**: Voltage stability constraints ensure that the power system can maintain acceptable voltage levels under normal and contingent conditions. These constraints are crucial for avoiding voltage collapse, which can lead to widespread blackouts [98-100].

**Transient Stability**: Transient stability constraints focus on the power system's ability to maintain synchronism following large disturbances, such as faults or sudden large load changes [101]. Incorporating these constraints helps in designing control actions that mitigate the risk of system instability or separation [102-103].

**Frequency Stability**: As the balance between supply and demand must be maintained at all times to keep the system frequency within a narrow range, frequency stability constraints are vital. These constraints are becoming increasingly important with the integration of variable renewable energy sources that can cause rapid frequency fluctuations [104-106].

**Small Signal Stability**: Small signal stability constraints address the system's ability to damp out small oscillations following minor disturbances. Ensuring small signal stability is essential for maintaining the overall system performance and preventing the escalation of oscillations into larger issues [107].

### 4.3.2 Related Research Findings and Applications

**Advanced Stability Modeling**: Recent research has focused on developing more sophisticated models for incorporating stability constraints into OPF. This includes dynamic models that more accurately represent the physical behaviors of the power system under different operating conditions [108].

**Integration with Renewable Energy**: With the variability and uncertainty introduced by renewable energy sources, new approaches to stability constraint formulation are being explored. These approaches consider the stochastic nature of wind and solar power generation, ensuring that stability is maintained even under highly variable conditions.

**Real-Time Stability Assessment**: Leveraging advancements in computing and data analytics, real-time stability assessment tools are being integrated into OPF models. These tools allow system operators to dynamically adjust operating strategies based on real-time data, enhancing system security and resilience [109-110].

**Decentralized and Distributed Approaches**: To address the distributed nature of modern power systems, research has also focused on decentralized and distributed methods for enforcing stability constraints. These methods allow for local control actions that can collectively ensure system-wide stability, offering scalability and flexibility in system operation.

**Stability-Constrained Market Operations**: In deregulated electricity markets, incorporating stability constraints into market operation models is vital for ensuring that the outcomes are not only economically efficient but also physically feasible. This involves developing market mechanisms and bidding strategies that account for the cost of maintaining stability [111].

In summary, incorporating security and stability constraints into OPF models is essential for the safe and reliable operation of modern power systems. Ongoing research in this area seeks to address the challenges posed by the increasing complexity of power systems, particularly with the integration of renewable energy sources, ensuring that these systems remain stable and resilient under a wide range of conditions.

## 4.4 Privacy-Preserving OPF

With the advent of smart grids and the increasing digitization of power systems, privacy concerns have become more prominent [112-114]. The OPF calculations, which are central to the operation of modern power systems, now often involve sensitive information that could be exploited if not properly protected [115]. This section delves into the necessity for privacy protection in OPF and explores potential solutions and case analyses demonstrating how these challenges can be addressed.

### 4.4.1 Necessity for Privacy Protection

**Consumer Data Sensitivity**: Smart meters and other IoT devices collect detailed consumption data that could reveal personal habits and behaviors of consumers. Protecting this information is crucial to maintaining consumer trust and complying with privacy regulations [116].

**Operational Security**: Detailed OPF data can potentially expose vulnerabilities in the power system's operation to malicious entities, posing risks to both physical and cyber security. Ensuring the privacy of this data helps in safeguarding the integrity and security of the power grid [117].

**Competitive Advantage**: In deregulated markets, the strategic information contained in OPF computations, such as generation capacities and bidding strategies, is valuable for maintaining competitive advantage. Privacy-preserving methods are necessary to prevent unfair competition and market manipulation.

#### 4.4.2 Solutions and Case Analyses

**Homomorphic Encryption**: This technique allows for operations to be performed on encrypted data, producing an encrypted result that, when decrypted, matches the result of operations performed on the plaintext. Case studies in OPF applications have demonstrated the feasibility of using homomorphic encryption to protect sensitive data during computation, albeit with challenges related to computational complexity and efficiency [118].

**Differential Privacy**: By adding controlled noise to the OPF data or outputs, differential privacy techniques ensure that the privacy of individual data points is preserved while still providing useful aggregated information. Case analyses in power systems have explored the balance between privacy levels and the accuracy of OPF solutions, showing that differential privacy can be effectively applied to protect consumer data [119].

**Secure Multi-Party Computation (SMPC)**: SMPC allows multiple parties to jointly compute a function over their inputs while keeping those inputs private. In the context of OPF, SMPC has been used to enable utilities and system operators to collaborate on power flow optimization without revealing sensitive operational or strategic information to each other. Case studies highlight the potential for SMPC to enhance privacy and security in collaborative optimization tasks, although scalability remains a challenge [120].

**Federated Learning**: This approach involves training machine learning models across multiple decentralized devices or servers holding local data samples, which is shown in Fig. 4. The fundamental prerequisite for employing a federated learning framework is to maintain the model's performance. In comparison to centralized training approaches, the performance standards for the federated learning method must satisfy specific criteria [121]:

$$\left|\psi_C - \psi_F\right| < \sigma \tag{6}$$

Here, $\psi_C$ represents the evaluation criteria for the centralized model, $\psi_F$ denotes the evaluation criteria for the federated model, and $\sigma$ is a small non-negative number. Applied to OPF, federated learning can help in developing predictive models and optimization algorithms that improve grid operation without compromising the privacy of individual data sources. Pilot projects in smart grids have begun to investigate the efficacy and practicality of federated learning for privacy-preserving analytics.

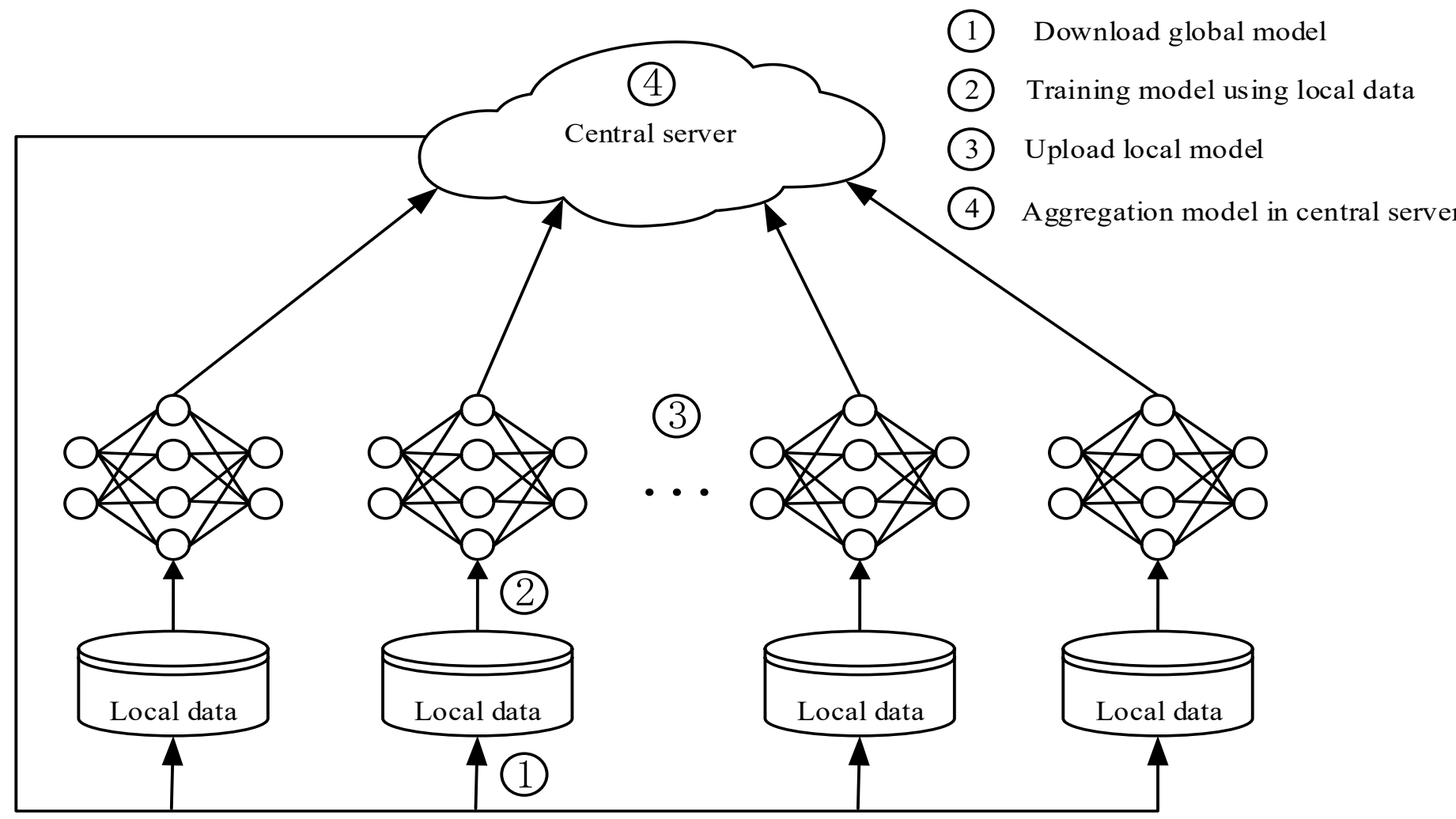

**Fig. 4 Principle of federated learning**

To summarize, the necessity for privacy protection in OPF is driven by the sensitivity of consumer data, the security of power system operations, and the competitive nature of energy markets. Various solutions, including homomorphic encryption, differential privacy, SMPC, and federated learning, offer promising paths to address these privacy concerns. Continued research and development, along with case studies and pilot implementations, are crucial for advancing these technologies and ensuring their effectiveness and scalability in real-world power systems.

### 4.5 OPF Based on Deep Reinforcement Learning

DRL combines the perception and generalization capabilities of deep learning with the decision-making ability of reinforcement learning [122-124]. This approach has shown significant promise in addressing the OPF problem, offering a dynamic and adaptive methodology capable of handling complex power systems with high-dimensional state and action spaces.

#### 4.5.1 Basic Principles of Deep Reinforcement Learning

**Reinforcement Learning (RL)**: At its core, RL involves an agent that learns to make decisions by interacting with an environment. Through this interaction, the agent receives feedback in the form of rewards or penalties, guiding it to learn the optimal policy for achieving its objectives.

**Deep Learning (DL)**: DL utilizes neural networks with multiple layers (deep networks) to learn hierarchical representations of data. When applied to RL, deep learning enables the agent to understand and interpret complex, high-dimensional inputs from its environment.

**DRL Framework**: In DRL, a deep neural network predicts the value of actions or directly derives the optimal policy that maximizes the cumulative reward over time [125]. This approach allows the RL agent to effectively process and act upon complex environmental states, such as those encountered in power systems.

In Deep Reinforcement Learning (DRL), agents seek to identify the optimal action strategy to maximize rewards through ongoing interactions with their environment. The outcomes of these interactions are determined solely by the current state, independent of past states. Consequently, this scenario is framed as a Markov Decision Process (MDP), characterized by four primary elements ($S$, $A$, $R$, $\pi$). The structure of the MDP is depicted in Fig. 5 [126].

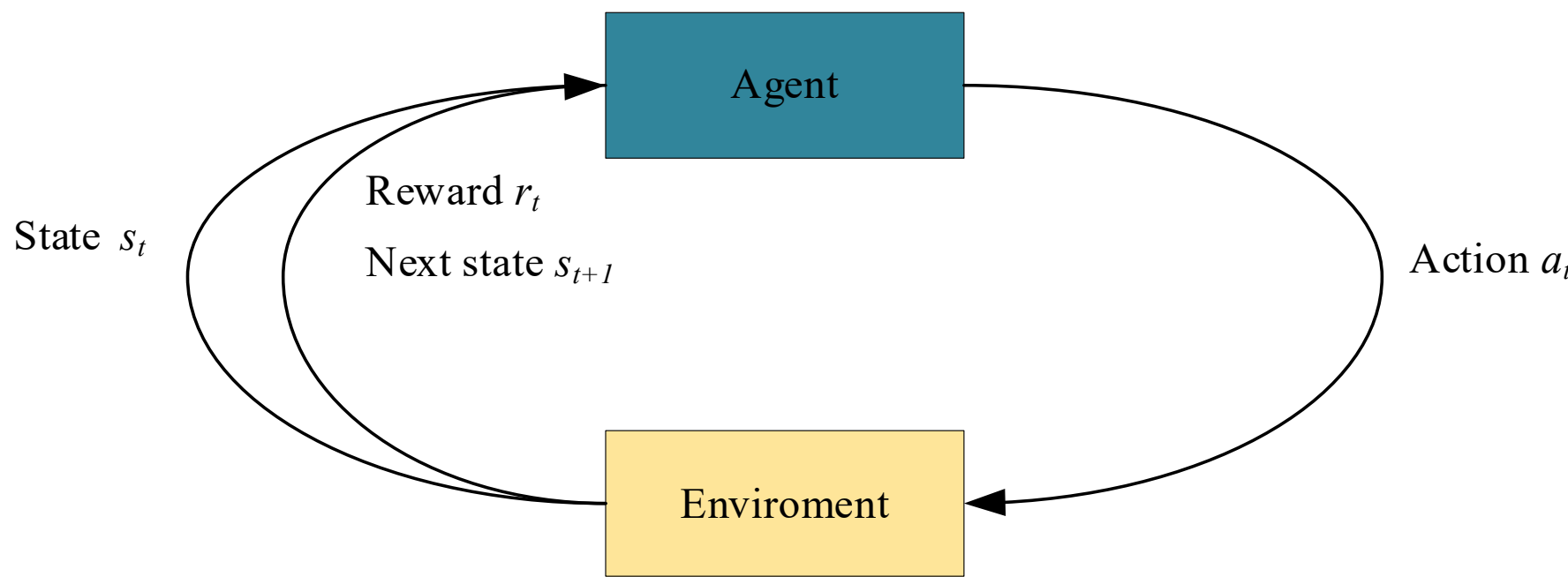

**Fig. 5. Markov decision process**

where $S$ is the environment state space, $A$ denotes the action space, $R$ stands for the reward function, $\pi$ denotes the policy.

### 4.5.2 Application Cases and Performance Evaluation

**Dynamic OPF**: DRL has been applied to dynamic OPF problems where the operational conditions of the power system change over time. In these cases, DRL agents have been trained to adjust generation dispatch and other control variables in real-time, optimizing operational objectives while maintaining system stability and adhering to constraints [127]. Performance evaluations have shown that DRL can outperform traditional optimization methods in terms of speed and adaptability to changing conditions.

**Integration of Renewable Energy**: DRL has also been explored for managing the integration of renewable energy sources into the grid. By dynamically optimizing power flow to accommodate the variability and uncertainty of renewables, DRL-based OPF strategies have demonstrated the potential to enhance grid reliability and reduce reliance on conventional generation sources [128]. Evaluations indicate that DRL can effectively balance supply and demand, minimize operational costs, and reduce emissions.

**Demand Response**: Another application case involves using DRL for optimizing demand response strategies within the OPF framework. DRL agents learn to modulate demand in response to grid conditions and market signals, achieving cost savings and improved grid stability [129]. Performance comparisons have highlighted the ability of DRL to generate efficient and responsive demand-side management strategies, offering significant benefits over static optimization approaches.

**Performance Evaluation Metrics**: The effectiveness of DRL in OPF applications is typically assessed using metrics such as cost reduction, emission reduction, improvement in grid stability, and the ability to maintain operational constraints [130]. Additionally, the computational efficiency and scalability of DRL approaches are critical factors, especially for real-time applications.

In summary, DRL presents a powerful tool for addressing the complexities of modern power systems, especially in dynamic and uncertain environments. Its applications in OPF range from dynamic system optimization to the integration of renewable resources and demand response management. Ongoing research and development efforts are focused on enhancing the performance, efficiency, and applicability of DRL-based OPF solutions, promising to revolutionize the way power systems are operated and managed.

### 4.6 OPF in Integrated Energy Systems (IES)

Integrated Energy Systems (IES) represent a holistic approach to energy management, combining electricity, heating, cooling, and other energy vectors to optimize efficiency, reliability, and sustainability [131-133]. These systems are increasingly recognized for their potential to facilitate the transition to a more sustainable energy future by leveraging synergies between different energy sectors [134,135]. For instance, electric-heat IES and electric-gas-heat IES are respectively shown in Fig. 6 [136] and Fig. 7 [137].

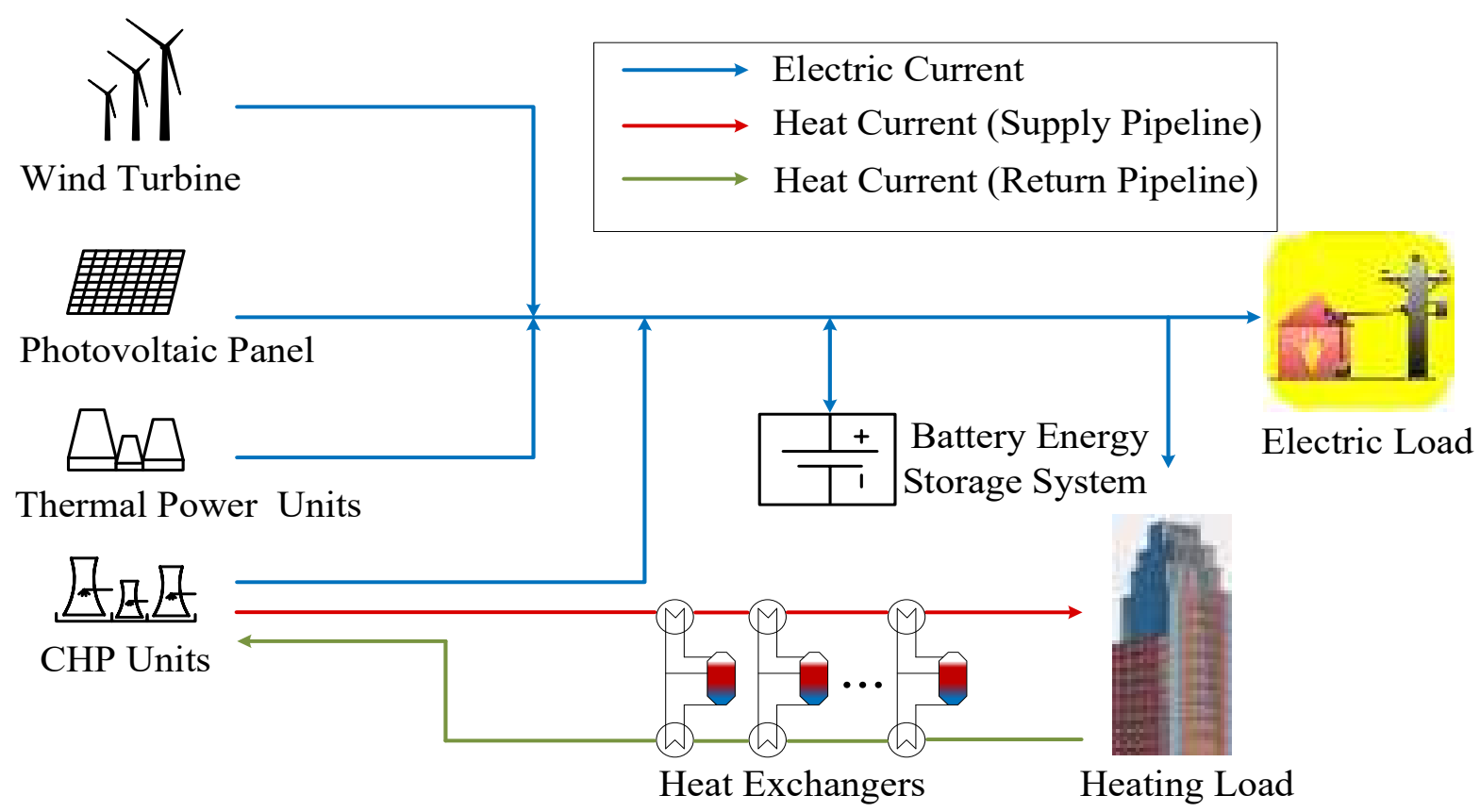

**Fig. 6. Schematic diagram of an electric-heat IES**

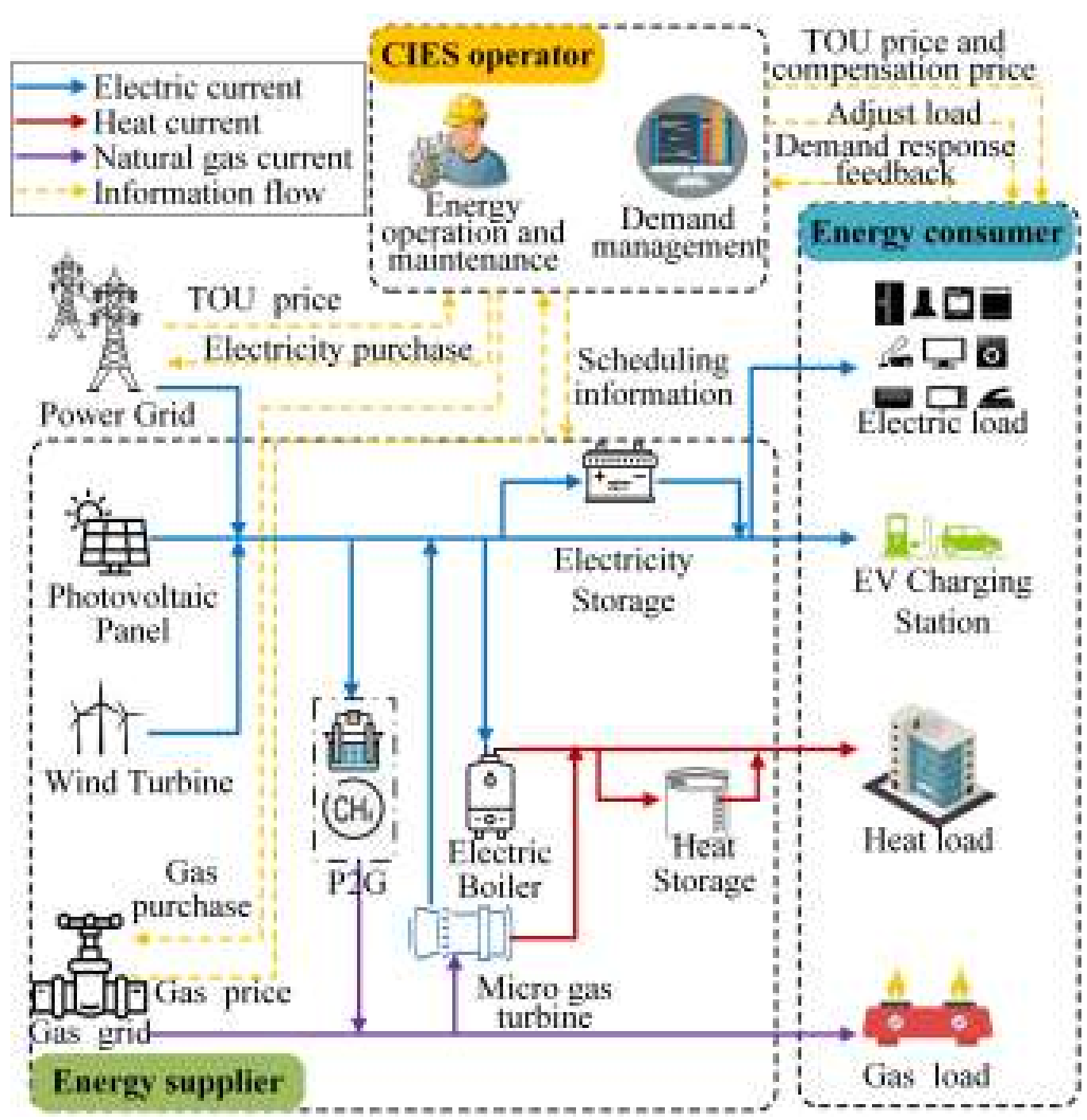

**Fig. 7. Schematic diagram of an electric-gas-heat IES**

### 4.6.1 Introduction to IES

**Concept and Components**: IES encompass a wide range of energy resources and infrastructures, including renewable energy sources, thermal energy storage, electric vehicles, and flexible loads, integrated through advanced control and optimization frameworks [138-139]. This integration allows for the exploitation of complementary energy dynamics, such as using excess electricity for heat production or utilizing waste heat to improve overall energy efficiency [140].

**Benefits and Challenges**: The primary benefits of IES include improved energy efficiency, reduced environmental impact, enhanced system resilience, and the ability to provide a more flexible response to varying energy demands [141]. However, the complexity of IES poses significant challenges for optimization, as it requires coordinating operations across diverse energy systems with different dynamics, constraints, and objectives.

### 4.6.2 OPF Issues and Solutions in IES Systems

**Complexity of Multi-Energy Flows**: Unlike traditional power systems, IES involve the simultaneous optimization of multiple energy carriers, each with its own set of technical and economic characteristics. This complexity requires advanced OPF models that can capture the interactions between different energy systems and optimize their joint operation [142].

**Integration of Renewable Energy**: The high penetration of renewable energy sources in IES introduces variability and uncertainty in energy supply, complicating the OPF problem [143]. Solutions have focused on developing stochastic and robust optimization techniques that can accommodate the uncertain nature of renewables and ensure reliable system operation under various scenarios.

**Demand-Side Management and Flexibility**: IES offer significant opportunities for demand-side management and the utilization of flexible loads to enhance system efficiency and stability. OPF solutions in IES have explored the use of dynamic pricing, demand response strategies, and energy storage to balance supply and demand, mitigate the impact of renewables' variability, and optimize overall system performance.

**Thermal and Electrical Coupling**: The coupling of thermal and electrical systems in IES introduces additional dimensions to the OPF problem, requiring models that can accurately represent the interdependencies between heat and power flows [144-145]. Solutions have included the development of integrated OPF models that simultaneously optimize both thermal and electrical systems, taking into account the potential for energy conversion and storage across different forms.

**Scalability and Computational Efficiency**: The increased complexity of IES and the need for high-resolution optimization across multiple energy carriers demand significant computational resources [146]. Research has focused on developing scalable optimization algorithms and leveraging advanced computing techniques, such as parallel computing and machine learning, to improve the computational efficiency of OPF in IES.

To summarize, OPF in Integrated Energy Systems presents a multifaceted challenge that spans technical, economic, and environmental dimensions. Addressing these challenges requires innovative optimization frameworks that can handle the complexity and interconnectivity of IES, paving the way for more sustainable and efficient energy systems. As research in this area progresses, it is expected that OPF solutions will increasingly enable the effective integration of diverse energy resources, enhancing the flexibility and resilience of the energy system as a whole.

### 4.7 OPF under Cyber Attacks

As power systems evolve with the integration of more digital technologies and the Internet of Things (IoT), they become increasingly susceptible to cyber-attacks [147-149]. These attacks can compromise the operational integrity and the security of the entire system [150-151]. This subsection will delve into the specifics of how Optimal Power Flow (OPF) systems can be targeted and affected by such threats, examine the adequacy of existing solutions, and explore emerging technologies that enhance resilience.

#### 4.7.1 Nature of Cyber Threats Specific to Power Systems

Cyber threats to power systems can vary widely, ranging from unauthorized access to sensitive data to direct attacks on the operational technology that controls power flows [152]. These threats include:

- **Denial of Service (DoS) Attacks**: These attacks can incapacitate power management systems, effectively shutting down operations and causing widespread disruptions [147].
- **Data Integrity Attacks**: These attacks compromise the accuracy and reliability of the operational data, leading to unsafe grid conditions. They include:
  - ✓ **False Data Injection Attacks (FDIA)**: This sophisticated type of data integrity attack involves injecting erroneous data into the system specifically, designed to evade detection by the system’s state estimation and its bad data detection functionality [153-154]. The aim is to manipulate decision-making processes, such as OPF, leading the system to make decisions that could destabilize the grid or result in inefficient power distribution.
  - ✓ **Other Forms of Data Tampering**: These may involve altering logged data, modifying command signals, or corrupting data files, all of which are aimed at causing operational disruptions or financial losses [155].
- **Man-in-the-Middle (MITM) Attacks**: These attacks occur when an attacker intercepts and manipulates the communication between different components of the power system, altering the information being exchanged [156].

These cyber threats not only jeopardize the operational efficiency of power systems but also pose significant financial risks and can erode trust among consumers and stakeholders [157-158].

#### 4.7.2 Existing Solutions and Their Limitations

While there are several cybersecurity measures in place, they often fall short in addressing the full spectrum of potential cyber threats faced by OPF systems. Current solutions include:

- **Firewalls and intrusion detection systems (IDS)** that monitor and control the incoming and outgoing network traffic.
- **Encryption techniques** that secure data in transit and at rest.

- **Regular security audits and compliance checks** to ensure that all system components meet the latest security standards.

However, these measures have limitations. For example, firewalls and IDS cannot prevent insider attacks or data manipulation by authenticated users [159]. Similarly, traditional encryption methods can introduce latency issues in real-time power system operations.

### 4.7.3 Innovative Approaches and Technological Advancements

To enhance the resilience of OPF to cyber attacks, the adoption of advanced cybersecurity technologies and methodologies is crucial. These include:

- **Blockchain technology** for secure, tamper-evident logging of all transactions and operations within the power system. This decentralized approach ensures that data integrity is maintained across all nodes in the network [160-161].

- **Advanced cryptographic techniques**, such as homomorphic encryption, which allows data to be processed while still encrypted, dramatically enhancing data security especially for cloud-based OPF applications.

- **AI and machine learning-based anomaly detection systems** that can continuously learn and adapt to new threats. These systems can detect subtle, unusual patterns in data that may indicate a cyber attack, often before traditional methods [162-164].

Additionally, **Zero Trust architectures**, where no entity is trusted by default from inside or outside the network, can significantly enhance the security posture of power systems by ensuring continuous verification and validation of all operations and data flows [165].

This subsection aims to provide a comprehensive analysis of how OPF systems can be safeguarded against cyber attacks, emphasizing the need for a dynamic and multifaceted approach to cybersecurity. By integrating these advanced technologies, the resilience of power grids against cyber threats can be substantially improved, ensuring reliable and secure power delivery in an increasingly digital age.

## 5. Technical Comparison and Analysis

The deployment of intelligent optimization algorithms in addressing the OPF problem, particularly in the context of modern power systems with integrated renewable energy sources and dynamic load demands, has been a focal point of recent research. This section provides a performance comparison of different intelligent optimization algorithms and a comparative analysis of their applications in real-world scenarios, highlighting the strengths, limitations, and suitability of each approach for various OPF challenges.

### 5.1 Performance Comparison of Different Intelligent Optimization Algorithms

**Genetic Algorithms**: GAs are well-suited for solving complex OPF problems due to their global search capabilities and flexibility in handling discrete and continuous variables. They excel in exploring large and complex search spaces but may require significant computational resources and time, especially for large-scale problems [166].

**Particle Swarm Optimization**: PSO is known for its simplicity, ease of implementation, and fast convergence rates [32]. It performs well in continuous search spaces and has been effective in OPF problems involving renewable energy integration. However, PSO can sometimes get trapped in local optima for highly complex or multimodal problems [103].

**Differential Evolution**: DE offers robustness and reliability in finding global optima and is effective in handling nonlinear, high-dimensional OPF problems. Its main advantage lies in its balance between exploration and exploitation. However, DE's performance can be sensitive to its parameter settings [166].

**Artificial Neural Networks** and **Deep Learning**: ANNs and DL models, particularly when combined with reinforcement learning, provide powerful tools for modeling and solving dynamic OPF problems. They excel in environments where the problem dynamics are highly nonlinear and uncertain [167-168]. The main challenge with these methods is the requirement for extensive training data and computational resources.

**Deep Reinforcement Learning**: DRL is advantageous for dynamic and sequential decision-making problems, making it suitable for real-time OPF applications. DRL can adapt to changing system conditions and learn optimal strategies over time [122]. However, designing and training DRL models is complex and computationally intensive [123].

### 5.2 Comparative Analysis of Real-World Application Cases

**Dynamic OPF in Smart Grids**: DRL has shown promising results in managing dynamic OPF challenges in smart grids, adjusting to real-time changes and optimizing power flow for reliability and efficiency. GAs and PSO are also applied but may not adapt as quickly to real-time changes as DRL.

**Renewable Energy Integration**: For OPF problems involving the integration of intermittent renewable energy sources, stochastic and robust versions of PSO and DE have been effectively used to account for the uncertainty and variability of renewables. DRL and hybrid algorithms combining GAs with PSO or DE have also demonstrated significant potential in optimizing the balance between renewable sources and traditional power generation.

**Demand Response Optimization**: PSO and GAs have been successfully applied to optimize demand response strategies within the OPF framework, providing solutions that enhance grid stability and reduce operation costs. DRL, with its ability to learn from environment interactions, offers a dynamic approach to demand response, potentially outperforming traditional optimization techniques in adapting to user behavior and market changes.

**Computational Efficiency and Scalability**: While GAs and PSO provide robust solutions for a wide range of OPF problems, their computational efficiency can be a concern for large-scale applications. DRL and DE, on the other hand, offer a good balance between solution quality and computational demands, especially when parallel computing techniques are employed.

To summarize, the selection of an intelligent optimization algorithm for OPF problems should be guided by the specific requirements of the application, including the problem's complexity, the need for real-time solutions, and the availability of computational resources. Hybrid approaches that combine the strengths of multiple algorithms and advanced models like DRL that can learn and adapt to changing conditions hold significant promise for addressing the evolving challenges of OPF in modern power systems.

## 6. Interdisciplinary Perspectives on MOPF Research

The evolution of MOPF research is increasingly characterized by its intersection with advances in computer science, artificial intelligence, data science, economics, and environmental science. This confluence has opened new avenues for addressing the complexities inherent in modern power systems, especially as they become more integrated, dynamic, and reliant on renewable energy sources. This section explores the latest developments in these fields and their application in MOPF research, highlighting the interdisciplinary methods that are shaping the future of energy systems optimization.

### 6.1 Latest Developments in Computer Science and Artificial Intelligence

**Machine Learning and Deep Learning**: The application of machine learning (ML) and deep learning (DL) techniques in MOPF research has led to significant breakthroughs in predictive modeling, anomaly detection, and real-time decision-making [169]. These approaches can dynamically adjust to changing system conditions, improving the accuracy of load and generation forecasts, and optimizing power flow in response to real-time data [170].

**Reinforcement Learning (RL)**: RL, particularly in its DRL variant, has emerged as a powerful tool for sequential decision-making problems in MOPF. DRL agents learn optimal policies through interactions with the environment, making it well-suited for dynamic and uncertain contexts like renewable energy integration and demand response management [171].

**Big Data Analytics**: The proliferation of data from smart grids and IoT devices has propelled the use of big data analytics in MOPF research. Advanced data processing and analytics techniques enable the extraction of meaningful insights from large datasets, facilitating more informed and effective optimization decisions [172].

**Blockchain and Distributed Ledger Technologies (DLT)**: Blockchain and DLT are being explored for their potential to enhance transparency, security, and decentralization in MOPF solutions. These technologies can enable secure, peer-to-peer energy transactions and decentralized energy resource management, paving the way for more resilient and flexible power systems [173].

### 6.2 Application of Interdisciplinary Methods in MOPF Research

**Integration of Economics and Renewable Energy**: Economic models help understand market dynamics and can be integrated with ML and optimization algorithms to manage the economic and technical aspects of renewable energy integration. This helps in developing strategies that not only meet technical requirements but also ensure economic viability and market compatibility.

**Environmental Impact and Optimization**: Environmental considerations are crucial as they provide a framework for assessing the sustainability of various MOPF strategies. By using AI to model environmental impacts, such as emission levels and resource consumption, MOPF research can contribute to more sustainable energy practices.

**Demand Response Optimization**: AI techniques, including ML and DRL, are applied to optimize demand response strategies within MOPF frameworks. Analyzing consumer behavior patterns and predicting load changes allow for more responsive and efficient demand-side management.

**Grid Stability and Security**: The integration of AI with traditional power system analysis tools enhances the ability to monitor, predict, and mitigate stability and security issues in real-time. For example, DL models can detect potential security threats or stability risks from high-dimensional data, allowing operators to take preemptive action.

In summary, the interdisciplinary fusion of MOPF research with advancements in computer science, AI, economics, and environmental sciences is driving the development of more intelligent, efficient, and sustainable power systems. By leveraging these cutting-edge technologies, researchers and practitioners can address the multifaceted challenges of modern energy systems, from integrating renewable energy sources to ensuring grid stability and engaging consumers in demand response programs. The continued exploration of these interdisciplinary methods promises to further revolutionize MOPF research and practice.

## 7. Standardization and Verification

In the evolving landscape of MOPF research, standardization and verification play crucial roles in ensuring the reliability, comparability, and reproducibility of optimization solutions. This section outlines the importance of standard test systems in MOPF research and discusses various methods for algorithm performance verification, highlighting how these practices contribute to the field's advancement.

### 7.1 Standard Test Systems in MOPF Research

**Role and Importance**: Standard test systems provide a common framework for evaluating and comparing the performance of different optimization algorithms under a set of predefined conditions. These systems typically represent simplified models of real-world power networks, including their buses, generators, loads, and transmission lines, along with specific operational constraints and objectives. By using these standardized models, researchers can objectively assess the effectiveness, efficiency, and robustness of their algorithms.

**Examples of Standard Test Systems**: Several test systems have become benchmarks in the field, such as the IEEE bus systems (e.g., IEEE 14, 30, 57, 118, and 300 bus test systems), which vary in complexity and size. More comprehensive test systems, like the MATPOWER package, offer a broader range of scenarios, including cases with renewable energy sources and storage options, to better reflect the challenges of modern power systems. For instance, IEEE RTS-79 test system is shown in Fig. 8 [174], and the IEEE 14-bus system with two- and three- terminal DC networks are illustrated in Fig. 9 and Fig. 10 [69].

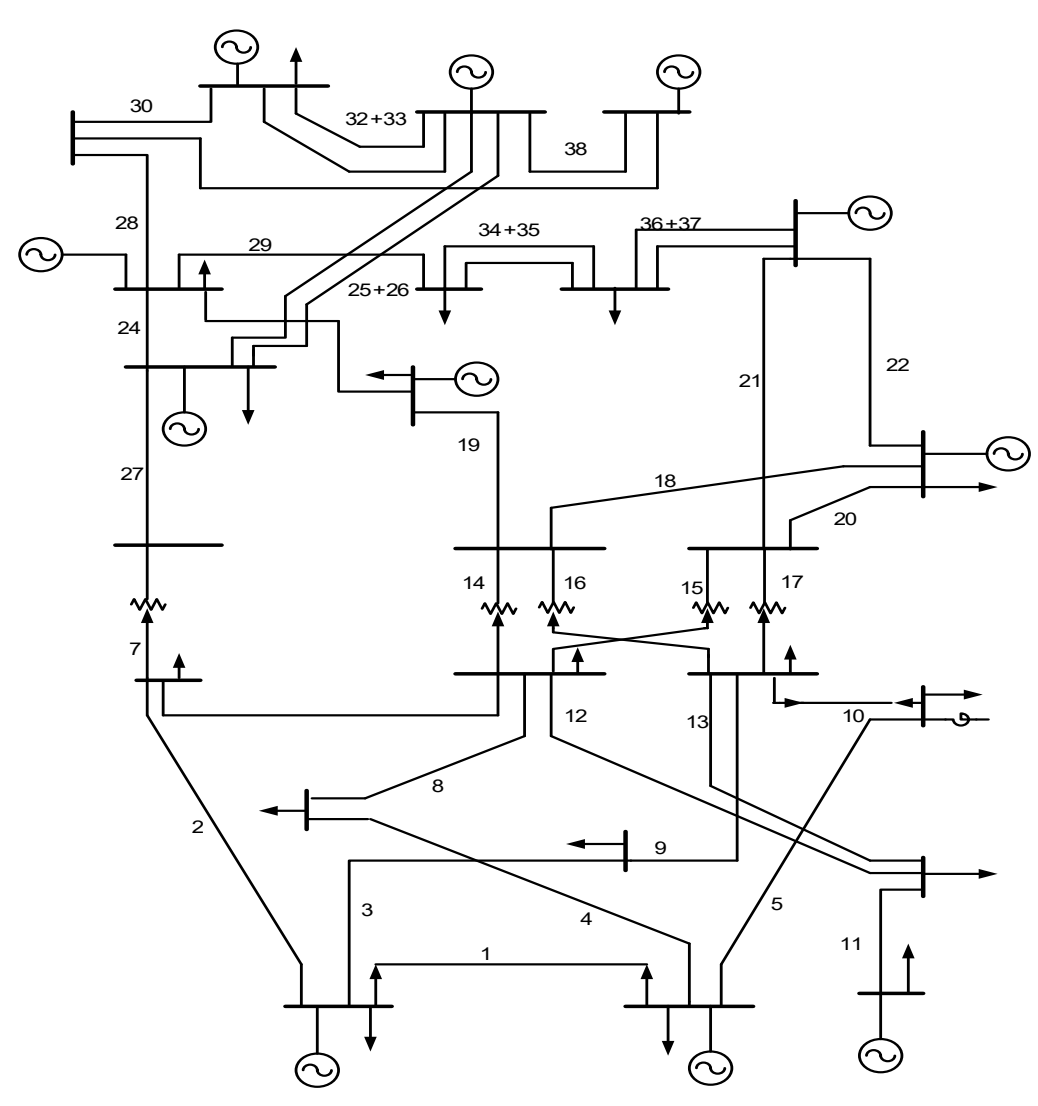

**Fig. 8 IEEE RTS-79 test system**

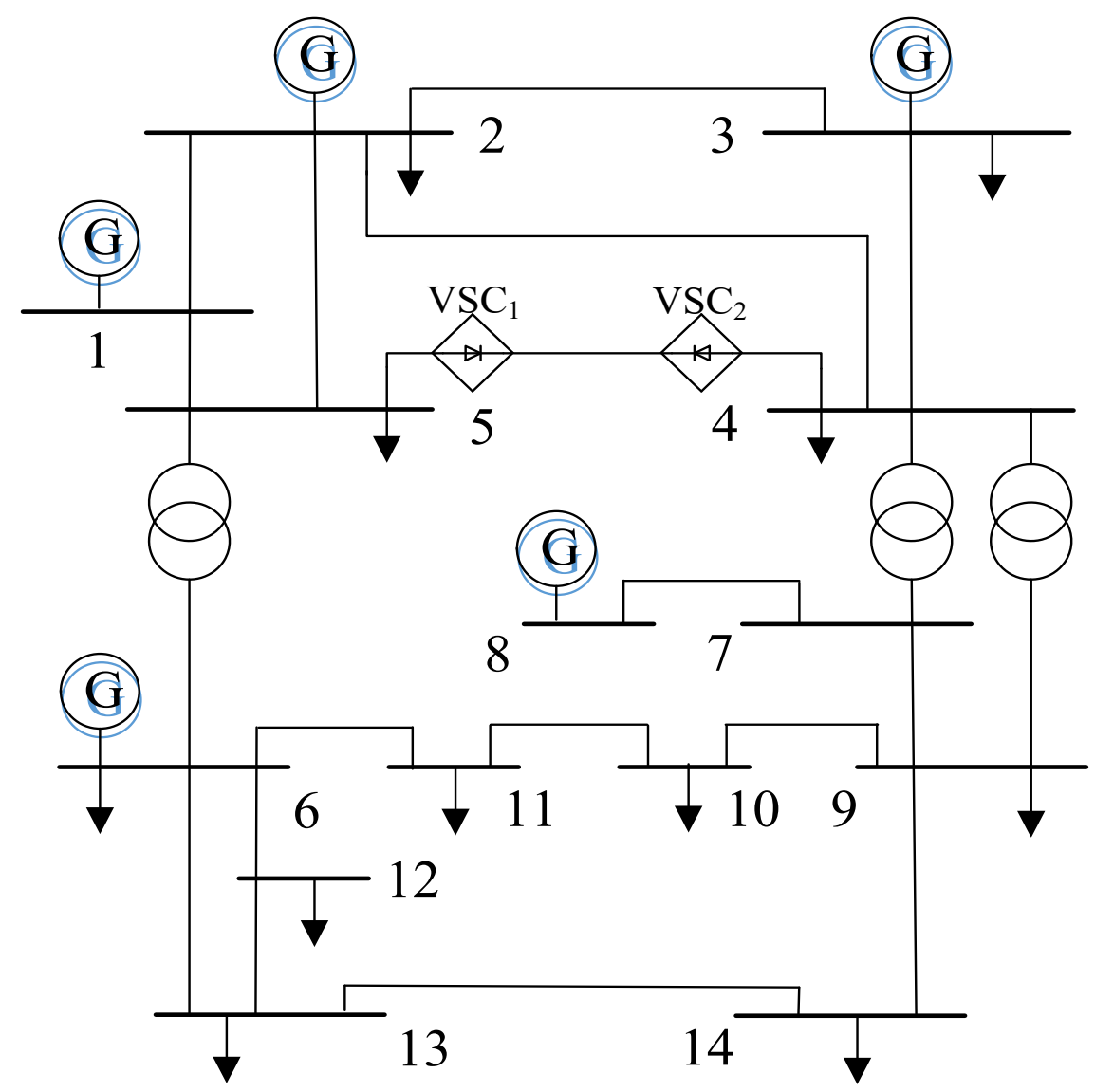

**Fig. 9.  IEEE 14-bus system with two-terminal DC network**

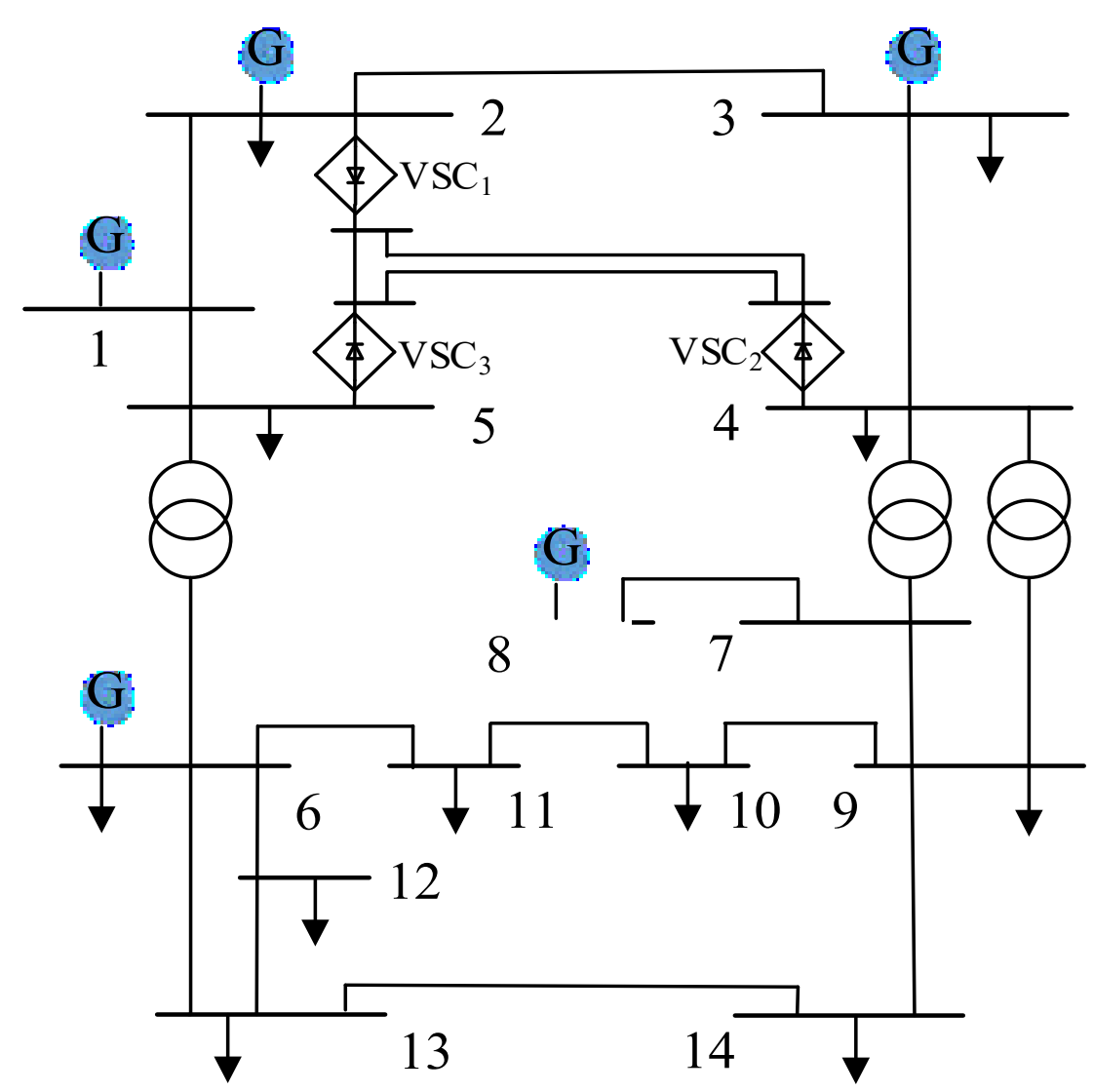

**Fig. 10.  IEEE 14-bus system with three-terminal DC network**

**Adaptation for MOPF**: While many standard test systems were initially designed for single-objective OPF, they have been adapted to support multi-objective optimization by incorporating additional objectives relevant to contemporary power systems, such as minimizing emissions, enhancing stability, or maximizing the integration of renewable energy sources.

**7.2 Methods for Algorithm Performance Verification**

**Benchmarking Against Standard Solutions**: One common method for verifying the performance of MOPF algorithms is benchmarking them against known solutions or the best results reported for standard test systems. This approach provides a direct comparison of solution quality and computational efficiency.

**Sensitivity Analysis**: Conducting sensitivity analyses to evaluate how changes in input parameters (such as demand levels, fuel costs, or renewable generation capacity) affect the algorithm's performance can offer insights into its robustness and adaptability to varying conditions.

**Statistical Testing**: Employing statistical tests to compare the results of different algorithms provides a rigorous method for assessing performance differences. Techniques such as t-tests, ANOVA, or non-parametric tests can determine whether observed differences in outcomes (like solution quality or computation time) are statistically significant.

**Cross-Validation in Real-World Scenarios**: Applying MOPF algorithms to real-world scenarios and cross-validating the results with actual operational data or expert assessments can verify the practical applicability and reliability of the solutions generated.

**Scalability Tests**: Evaluating how an algorithm performs as the size and complexity of the problem increase is crucial for understanding its scalability. Tests can gradually increase the number of buses, generators, and constraints in the system to assess the impact on computational resources and solution time.

To summarize, the standardization and verification of MOPF algorithms are foundational to the field's progress. Standard test systems offer a basis for consistent, comparative analysis, while diverse verification methods ensure the reliability, efficiency, and practical relevance of optimization solutions. As MOPF research continues to evolve, the development of new standards and verification methodologies will be vital for addressing the complexities of integrating emerging technologies and renewable energy sources into power systems.

## 8. Software and Tools Related to MOPF

The complexity of MOPF problems necessitates the use of sophisticated software and tools. These resources range from specialized optimization frameworks to comprehensive simulation platforms, facilitating the analysis, modeling, and optimization of power systems under various operational and environmental conditions. This section provides an overview of notable software and tools in the MOPF domain, along with guidelines and recommendations for their selection and use.

### 8.1 Introduction to Software and Tools

**MATPOWER**: A MATLAB-based power system simulation package designed for solving OPF problems, including MOPF scenarios [175]. It supports a wide range of standard test systems and offers functionalities for static and dynamic analysis, making it suitable for both research and educational purposes.

**PSS®E**: Developed by Siemens, PSS®E is a comprehensive tool for the analysis and optimization of power system operations. It includes advanced capabilities for load flow analysis, stability assessment, and OPF, accommodating large-scale and complex power system models [176].

**GridLAB-D**: An open-source distribution system simulation and analysis tool that provides insights into the operational dynamics of power systems with integrated renewable energy sources, demand response programs, and distributed generation. It is particularly useful for studying the impacts of distributed energy resources on grid operations [177].

**OpenDSS**: The Electric Power Research Institute's (EPRI) open-source distribution system simulator supports detailed modeling of distributed resources, including their impact on power flow and voltage profiles. It is well-suited for researchers focusing on the distribution side of power systems [178].

**hynet**: A Python-based open-source framework designed for the simulation and optimization of hybrid AC/DC power systems. It enables comprehensive analysis and operation planning with its robust support for convex relaxation techniques and a flexible software design tailored for both research and practical applications [179].

PowerModelsADA: An open-source tool built to facilitate the solving of Optimal Power Flow (OPF) problems through Alternating Distributed Algorithms (ADA). It offers a versatile environment for experimenting, validating, and comparing various ADA approaches, ensuring robust and efficient power system optimization [180].

### 8.2 Guidelines and Recommendations for Selection and Use

**Assess Compatibility and Requirements**: Before selecting a tool, evaluate its compatibility with your existing computational environment and its suitability for your specific research needs. Consider factors such as the scale of the power systems you aim to model, the complexity of the optimization problems, and the specific objectives of your MOPF studies.

**Explore Documentation and Community Support**: Comprehensive documentation and active user communities can significantly enhance your ability to effectively utilize a tool. Look for resources such as user manuals, tutorials, and forums where users discuss issues and share insights.

**Consider Open Source vs. Commercial Options**: Open-source tools offer the advantage of customization and no licensing fees, which can be beneficial for academic research and projects with limited budgets. Commercial software, however, often provides more extensive support and advanced features, which might be necessary for complex or large-scale projects.

**Evaluate Learning Curve and Usability**: Some tools may require significant time to learn and master, especially for users without extensive programming or power system analysis experience. Assess the usability and learning curve of the software, and consider the availability of training resources or courses.

**Future-Proofing Your Choice**: Consider the tool's adaptability to future research directions and its ability to handle emerging challenges in power systems, such as the integration of novel renewable energy technologies or the modeling of smart grid functionalities.

In summary, the selection of software and tools for MOPF research should be guided by a careful evaluation of your project's specific needs, the tool's features and capabilities, and the support resources available. Balancing these factors will help ensure that you choose the most appropriate and effective tools for your work, enabling you to conduct thorough and impactful research in the domain of MOPF.

## 9. Conclusion

The exploration of MOPF within the context of modern power systems underscores the field's dynamic evolution and its critical role in navigating the complexities of today's energy challenges. This journey through MOPF research highlights the intersection of advanced optimization techniques, interdisciplinary methods, and innovative technologies, all aimed at enhancing the efficiency, reliability, and sustainability of power systems. As we conclude, we summarize the key findings from this comprehensive review and look ahead to future research directions and the outlook for the field.

**9.1 Summary of Research Findings**

- **Intelligent Optimization Algorithms**: The application of sophisticated algorithms, including genetic algorithms, particle swarm optimization, and deep reinforcement learning, has significantly advanced MOPF solutions, offering enhanced capability to handle the complexity and multi-dimensionality of modern power systems.
- **Integration of Renewable Energy Sources**: Addressing the variability and uncertainty of renewable energy within MOPF frameworks has been a focal point, with stochastic and robust optimization methods proving effective in ensuring system stability and efficiency.
- **Interdisciplinary Approaches**: The integration of computer science and artificial intelligence with power system engineering has opened new avenues for predictive modeling, real-time decision-making, and the management of distributed energy resources.
- **Standardization and Verification**: The adoption of standard test systems and rigorous verification methods has been instrumental in ensuring the reliability and comparability of MOPF solutions, fostering a culture of transparency and scientific rigor in the field.
- **Software and Tools**: A diverse range of software and tools has facilitated the modeling, analysis, and optimization of power systems, supporting both academic research and practical applications in the energy sector.

**9.2 Future Research Directions and Outlook**

- **Advanced Machine Learning Models**: The potential of machine learning, especially deep learning and reinforcement learning, to further refine MOPF solutions is immense. Similarly, deep learning methods, such as BiLSTM-GCN, have shown remarkable capability in feature mining and recognition tasks, suggesting potential applications in power systems for dynamic optimization and real-time monitoring [181]. Future research will likely focus on exploiting these models to better predict system dynamics, optimize in real-time, and handle the stochastic nature of renewables.
- **Decentralization and Blockchain**: As power systems move towards greater decentralization, research into blockchain and distributed ledger technologies for secure, transparent, and efficient energy transactions and grid management will become increasingly important.
- **Multi-Energy Systems Optimization**: The optimization of integrated energy systems, incorporating electricity, heat, gas, and other energy vectors, presents a fertile ground for future MOPF research, aiming to unlock synergies across different energy sectors.
- **Climate Resilience and Adaptation**: Developing MOPF strategies that enhance the resilience of power systems to climate change impacts and extreme weather events will be crucial, ensuring reliable energy supply in the face of increasing environmental uncertainties.

- **Policy and Regulatory Frameworks**: Bridging the gap between technological advancements and policy development will be essential. Future research should also explore the implications of MOPF solutions within regulatory frameworks, market structures, and energy policies to facilitate their adoption and impact.

In conclusion, the field of MOPF research is poised for significant advancements as it embraces new technologies, confronts emerging challenges, and seeks to harmonize the objectives of efficiency, reliability, sustainability, and resilience. The journey ahead is both exciting and demanding, requiring a concerted effort from researchers, practitioners, policymakers, and stakeholders to navigate the evolving energy landscape and contribute to a sustainable future.